\documentclass[10pt,twocolumn,letterpaper]{article}

\usepackage{wacv}
\usepackage{times}
\usepackage{epsfig}
\usepackage{graphicx}
\usepackage{amsmath}
\usepackage{amssymb}
\usepackage{bm}
\usepackage[olditem,oldenum]{paralist}

\newcommand{\vx}{\bm{x}}
\newcommand{\vy}{\bm{y}}
\newcommand{\ve}{\bm{e}}

\newcommand{\vw}{\bm{w}}
\newcommand{\mW}{\bm{W}}

\newcommand{\seq}{\mathcal{S}}
\newcommand{\capt}{\mathcal{Y}}

\newcommand{\bias}{\bm{b}}
\newcommand{\todo}[1]{}
\newcommand{\enorm}[1]{\left\|{#1}\right\|_2}

\DeclareMathOperator*{\argmin}{arg\,min}
\newcommand{\half}{\frac{1}{2}}

\DeclareMathOperator*{\lstm}{LSTM}

\DeclareMathOperator{\softmax}{softmax}
\DeclareMathOperator{\softplus}{softplus}
\newcommand{\set}[1]{\left\{#1\right\}}
\newcommand{\simplex}[1]{\Delta^{#1}}
\DeclareMathOperator{\att}{att}

\newcommand{\reals}[1]{\mathbb{R}^{#1}}
\newcommand{\binary}[1]{\mathbb{B}^{#1}}

\wacvfinalcopy 

\def\wacvPaperID{1091} 

\ifwacvfinal\fi
\begin{document}

\title{Spatio-Temporal Ranked-Attention Networks for Video Captioning}

\author{
  Anoop Cherian$^1$\quad Jue Wang$^2$ \quad Chiori Hori$^1$ \quad Tim K. Marks$^1$\\
  $^1$Mitsubishi Electric Research Labs, Cambridge, MA\quad 
  $^2$Australian National University, Canberra\\
  {\tt\small \{cherian,chori,tmarks\}@merl.com\quad jue.wang@anu.edu.au}
}

\maketitle
\ifwacvfinal\thispagestyle{empty}\fi

\begin{abstract}
Generating video descriptions automatically is a challenging task that involves a complex interplay between spatio-temporal visual features and language models. Given that videos consist of spatial (frame-level) features and their temporal evolutions, an effective captioning model should be able to attend to these different cues selectively. To this end, we propose a Spatio-Temporal and Temporo-Spatial (STaTS) attention model which, conditioned on the language state, hierarchically combines spatial and temporal attention to videos in two different orders: (i) a spatio-temporal (ST) sub-model, which first attends to regions that have temporal evolution, then temporally pools the features from these regions; and  (ii) a temporo-spatial (TS) sub-model, which first decides a single frame to attend to, then applies spatial attention within that frame. We propose a novel LSTM-based temporal ranking function, which we call ranked attention, for the ST model to capture action dynamics. Our entire framework is trained end-to-end. We provide experiments on two benchmark datasets: MSVD and MSR-VTT. Our results demonstrate the synergy between the ST and TS modules, outperforming recent state-of-the-art methods.
\end{abstract}
\section{Introduction}
\label{sec:intro}
The recent advances enabled by deep neural networks in computer vision, audio, and natural language processing have stimulated researchers to look beyond these as isolated domains, instead tackling problems at their intersections~\cite{thomason2014integrating,ephrat2018looking,zhao2018sound,chen2015microsoft}. Automatic video captioning is one such multimodal inference problem that has gained attention in recent years~\cite{hori2017attention,wang2018video,wang2018watch}, thanks to the availability of sophisticated CNN models~\cite{carreira2017quo,feichtenhofer2017spatiotemporal,aytar2016soundnet,sutskever2014sequence} and massive training datasets for video activity recognition~\cite{kay2017kinetics,gu2017ava,johnson2016densecap}, audio classification~\cite{gemmeke2017audio}, and neural machine translation~\cite{pennington2014glove,bahdanau2014neural}. However, learning to describe video data is still a challenging problem, as generating good captions requires inferring the intricate relationships and interactions between subjects and objects in a video. Despite recent progress~\cite{chen2018less,hori2017attention,wang2018video,wang2018watch}, this task remains difficult. This may be due to the high dimensionality of spatio-temporal data, which can generate large volumes of features of which only a few may be correlated to the way humans describe videos.

Taking inspiration from neural translation models, one promising way to approach the video captioning problem is to leverage~\emph{visual attention}~\cite{tu2017video,zanfir2016spatio,anderson_updown,xu2017learning}. Such techniques use the compositional nature of language models to attend to specific visual cues in order to generate subsequent words in a caption. Attention has also been explored for multimodal fusion using image, audio, and motion cues~\cite{wang2018watch,hori2017attention}. However, these works consider frame-level or clip-level representations of videos, which may not capture specific details of the scene or may represent too much information that is unrelated to the primary content.  
\begin{figure}
    \centering
    \includegraphics[width=8cm,trim={0.9cm 5.5cm 13cm 3cm},clip]{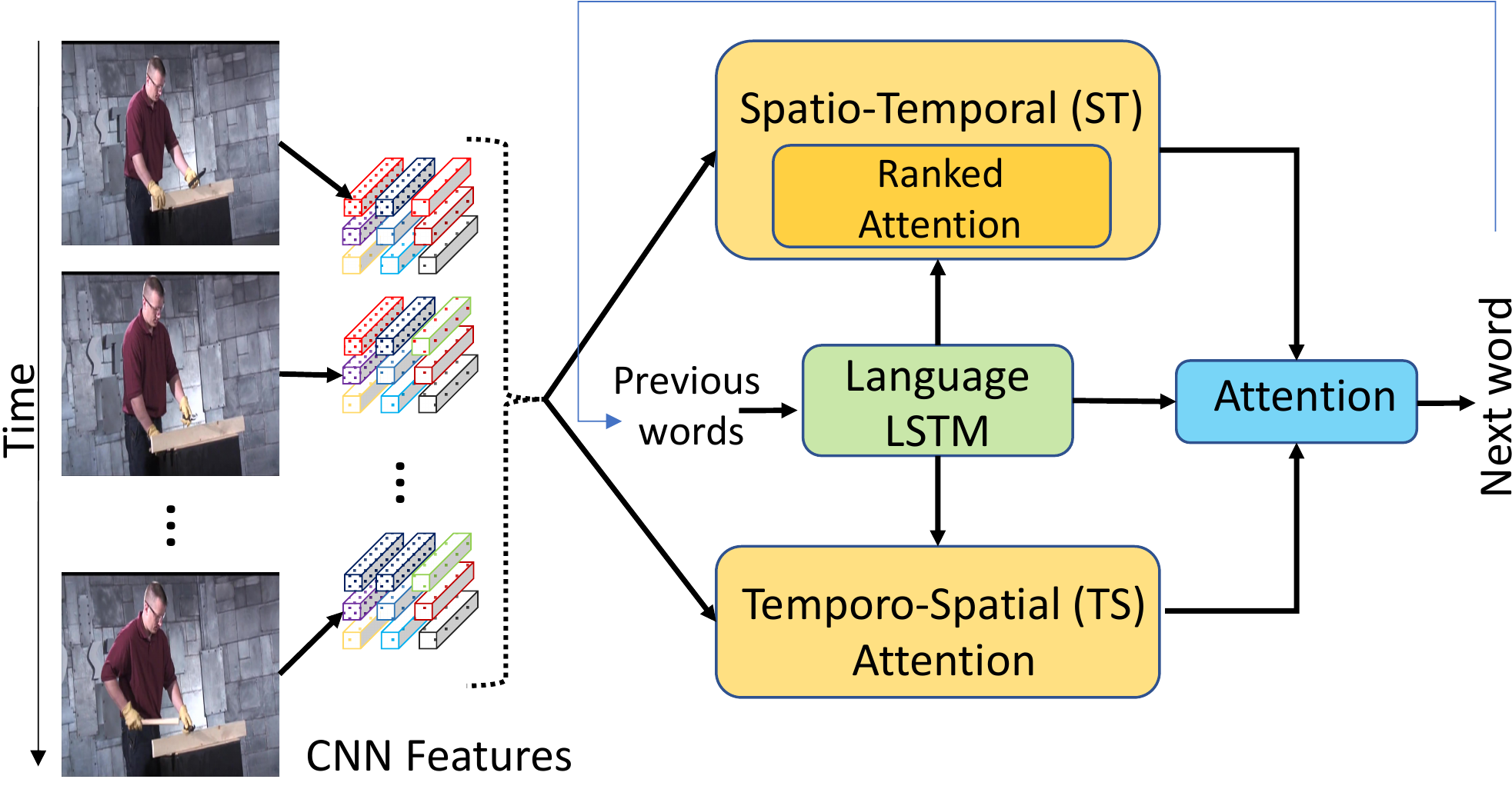}
    \caption{Our overall spatio-temporal and temporo-spatial (STaTS) attention architecture.}
    \label{fig:overall_arch}
\end{figure}

There have been efforts to address such granularity issues by using spatial attention, as for example in
image captioning~\cite{anderson_updown,yang2017catching}. Such schemes usually use a pre-trained object detector, e.g., Fast RCNN~\cite{ren2015faster}, which may be useful for detecting specific objects in the scene but may miss out on the scene context or visual cues related to human actions or interactions. One could also use schemes such as action proposals~\cite{krishna2017dense,yu2015fast}, but they can be computationally expensive. This paper is similar in vein to these works, in that we also explore video captioning using spatial and temporal attention. However, we apply and combine these attentions in a novel way.

Our main contribution is an attention model that we call STaTS (Spatio-Temporal and Temporo-Spatial). Our model, illustrated in Figure~\ref{fig:overall_arch}, hierarchically combines spatial and temporal attention in two different orders, which we call spatio-temporal (ST) attention and temporo-spatial (TS) attention. For ST attention, we first apply spatial attention and linear pooling on deep features derived from each video frame, then apply a temporal attention over these features.  The ST model's composition of spatial and temporal attention modules helps reduce the size of the spatial/temporal attention space from multiplicative to additive. 

Further, to ensure that temporal pooling captures the dynamic nature of actions in videos, we introduce a novel LSTM-based ranking formulation that attends to consecutive pairs of frames in a way that preserves their temporal order. We call this \emph{ranked attention}. Our key idea is to use an LSTM to emulate a rank-SVM~\cite{fernando2015modeling} such that the representation this module generates captures the temporal evolution of video features. Such a technique avoids the otherwise computationally challenging implicit differentiation that one needs to use for rank-pooling~\cite{fernando2016discriminative,gould2016differentiating}. 

One weakness of the ST model may be that not all words in a caption rely on such temporally varying holistic features. Words for the \emph{subject} or \emph{object}, for example, might be more directly obtained by considering more localized features from a single representative frame. To this end, we propose a novel temporo-spatial (TS) attention model that provides a shortcut for visual relationship inference, without going through the ST pipeline described above. Specifically, the TS pipeline first applies temporal attention to frame-level representations to (softly) select specific frames to attend to, then applies spatial attention to the spatial feature representations of these frames. 

Our STaTS model generates two attention-weighted video representations (ST and TS), which we combine via a weighted average, conditioned on the state of the language model (sentence generator), where these weights are computed by passing the two representations through a further attention scheme across the ST and TS models.

In Section~\ref{sec:expts}, we present experiments evaluating the benefits of each of the above modules. We base our experiments on two frequently used video captioning benchmarks: the MSVD (YouTube2Text)~\cite{guadarrama2013youtube2text} and MSR-VTT~\cite{xu2016msr} datasets. For the spatial features, we explore the advantages of using 3D CNN features from the recent Inflated 3D (I3D) activity recognition model~\cite{carreira2017quo}, as well as features from a Fast RCNN object detection model~\cite{ren2015faster}. Our experiments clearly demonstrate the advantages of our STaTS model, leading to state-of-the-art results on the MSVD dataset on all evaluation metrics. On MSR-VTT, we achieve the best performance on some metrics and are competitive with the recent state of the art on others.

We now summarize the main contributions of this work:
\begin{compactenum}
\item We present a novel spatio-temporal and temporo-spatial attention model, in which each of the two sub-models selectively attends to complimentary visual cues required to generate sentences.
\item We propose a novel temporal attention scheme, {\em ranked attention}, by formulating an LSTM-based objective that emulates a rank-SVM algorithm for temporally ordered feature aggregation.
\item We present extensive experiments and analysis on two benchmark datasets, using varied 2D and 3D CNN-based feature representations, and demonstrate state-of-the-art performance. 
\end{compactenum}

\section{Related Work}
\label{sec:related_work}
\textbf{Video Captioning.} Traditional methods for video captioning are usually based on predefined language templates~\cite{guadarrama2013youtube2text,kojima2002natural,rohrbach2013translating,krishnamoorthy2013generating,thomason2014integrating,yu2013grounded,das2013thousand,hospedales2009markov,wanke2010topic}, which reduce a free-form caption generation model into one of recognizing the categories to fill in for various attributes and keywords in the template (such as the subject, verb, and object). For example, in Rohrbach et al.~\cite{rohrbach2013translating}, a conditional random field is proposed to model the correlation between activities and objects in the video. Markov models are also adopted to produce semantic features for sentence generation~\cite{yu2013grounded,das2013thousand,hospedales2009markov,wanke2010topic}. Such models disentangle the need for the language model to learn grammar, thereby simplifying the problem. However, the captions generated are limited by the syntactical structure, which limits their diversity and the system's ability to generalize. In contrast to these prior works, there have been recent efforts at leveraging deep recurrent architectures such as long short-term memory (LSTM) for sequence learning tasks, starting with the seminal work of Karpathy et al.~\cite{karpathy2015deep}. Venugopalan et al.~\cite{venugopalan2014translating} propose an LSTM-based model to generate captions from temporally average-pooled CNN visual features. Since the average pooling destroys the temporal dynamics of the sequence, Yao et al.~\cite{yao2015describing} present a temporal attention mechanism to associate a weighting for the feature from each frame and fuse them using a weighted average. Along similar lines, Venugopalan et al.~\cite{venugopalan2015sequence} introduce S2VT, which utilizes LSTMs in both encoder and decoder and includes optical flow to incorporate temporal dynamics. Zhang et al.~\cite{zhang2016automatic} propose a two-stream feature encoder to aggregate both spatial and temporal cues jointly using 3D CNN features. Hori et al.~\cite{hori2017attention} extend temporal attention by attending to different input modalities such as image, motion, and audio features. Our method differs from these in the way we disentangle the video features. Our approach allows simultaneous hierarchical and coupled extraction of  spatio-temporal video cues in a simple framework.

\textbf{Spatio-Temporal Attention.} As mentioned above, temporal attention has been widely used in recent video captioning work to decide which frame(s) in the video are important for generating the next word in a caption. However, these systems usually map the raw video frames into high-level CNN features (via a suitable spatial pooling operator), which marginalizes away important spatial information (such as location and class of specific objects or actions) that are important for captioning. 

Spatial-temporal video feature learning has been widely used in several video applications, such as video classification~\cite{peng2018two,feichtenhofer2016spatiotemporal,yu2017cascaded} and video super-resolution~\cite{yang2017video}. Related work in image captioning includes~\cite{anderson_updown}, which applies top-down and bottom-up attention to Fast R-CNN features, and~\cite{lu2017knowing}, which applies an attention-based LSTM to generate a spatially weighted feature map.
In video captioning, Yang et al.~\cite{yang2017catching} localize regions of interest in every frame using attention; however, not every frame may have have such a region, and they need additional semantic supervision to attend to informative regions. Zanfir et al.~\cite{zanfir2016spatio} propose a spatial-temporal attention model that assigns a weighting to both spatial and temporal CNN visual features from optical flow, RGB frames, and detected objects in each frame. Tu et al.~\cite{tu2017video} and Yu et al.~\cite{yu2016video} propose hierarchical attention schemes that condition on the current caption word and visual features. They first generate spatial attention weights, conditioned on which a similar attention scheme is adopted temporally; the weighted features are used to generate the word. While this scheme shares a similar motivation to ours, their attention model must select from a much larger number of features---a harder attention problem that demands larger datasets for training. We avoid this difficulty by attending to spatial and temporal features in stages, each stage reducing the data complexity. More recently, Aafaq et al.~\cite{aafaq2019spatio} use spatio-temporal feature engineering to improve captioning performance. In~\cite{Zhang_2019_CVPR}, object saliency is combined with bidirectional temporal graph reasoning; this is related to our proposed ranked attention model, but our formulation is much simpler.

\textbf{Reinforcement Learning (RL).} There are two key ways a video captioning problem can be cast in an RL setting: (i) selecting informative features or frames, and (ii) optimizing the training on evaluation metrics that are usually not differentiable (such as BLEU, CIDER, METEOR, etc.). For the former setting, several recent works have achieved promising results~\cite{chen2018less,wang2018video-rl} by picking suitable frames to encode based on a predesigned reward function. Chen et al.~\cite{chen2018less} incorporate visual diversity and CIDER score into the reward function. Similarly, \cite{wang2018video-rl} models a manager and a worker within a hierarchical LSTM  to achieve better feature encoding. When using RL to optimize non-differentiable losses, prior works typically use the policy-gradient algorithm~\cite{chen2018less}. While we believe our sophisticated attention scheme can pick visual features without needing an RL engine, we do use policy gradients to optimize our model for losses defined over METEOR and BLEU metrics (as in~\cite{rennie2017self}).
\section{Proposed Method}
\label{sec:method}
\begin{figure*}[t]
    \centering
    \includegraphics[width=16cm,trim={1.7cm 5.5cm 1.3cm 3cm},clip]{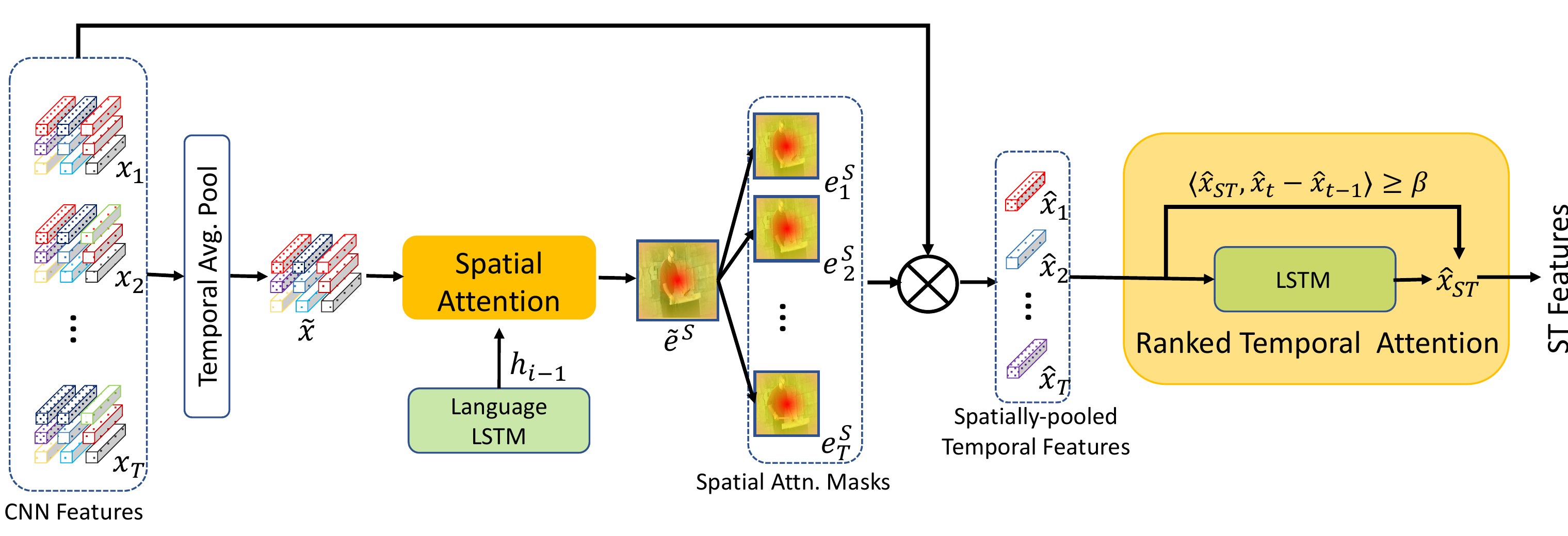}
    \caption{Our spatio-temporal (ST) network with the ranked temporal attention module.}
    \label{fig:st_attn_model}
\end{figure*}
In this section, we introduce our Spatio-Temporal and Temporo-Spatial (STaTS) attention model for video captioning, illustrated in Figure~\ref{fig:overall_arch}. In Section~\ref{sec:STspatial}, we describe our spatio-temporal (ST) attention model, which consists of a spatial attention model (Section~\ref{sec:STspatialAttention}) followed by our proposed ranked temporal attention model (Section~\ref{sec:STtemporalRanked}). We explain our temporo-spatial (TS) model in Section~\ref{sec:TS}. Finally, we describe how the ST and TS models are combined into our full STaTS attention model in Section~\ref{sec:STaTS}.

Before proceeding, let us review our notation. Suppose we are given a training set of $N$ videos,  \mbox{$\mathfrak{S}=\set{(\seq_1,\capt_1), (\seq_2,\capt_2), \cdots, (\seq_N, \capt_N)}$.} Here, $\seq_k$ is a temporally ordered sequence of frame-level features for video~$k$, and each $\capt_k$ is a textual description of the video (caption), the words of which are encoded using their indices in a predefined dictionary. Let each video sequence $\seq_k=\langle \vx_1, \vx_2, \cdots, \vx_T\rangle$ be a sequence of $T$ temporally ordered video frames. For each video frame $t$, we have $n$ features, denoted $\vx_{tj}$ for $t=1,2,\ldots, T$ and $j=1,2,\ldots, n$, where each $\vx_{tj}\in\reals{d}$. For each $j$, $\vx_{tj}$ encodes visual information from a different region (out of $n$ regions of the image). Such spatial features could be produced, for example, from each cell of a non-overlapping grid as from the intermediate spatial pooling layers of a CNN, or regions  obtained from an RCNN object detector. To encode captions, we assume each $\capt_k=\langle \vy_1, \vy_2, \cdots, \vy_{m}\rangle$ is an ordered sequence of word embeddings, where the $i$th word in the caption, $\vy_i \in \binary{D},$ is a one-hot vector encoded using a language dictionary of size~$D$.

Given that the size of the language dictionary $D$ is usually enormous, learning a neural network model to generate a caption with $m$ words would demand exploring a space of $D^m$ sentences, which may be computationally challenging. Fortunately, however, the language model is highly structured and compositional, so one can generate each word sequentially conditioned on the previously generated words. This idea is usually implemented via a long short-term memory (LSTM), which takes as input the current word $\vy_i$ in a sentence $\capt_k$ and a state representation $h_{i-1}$ of the previous words in the sentence, and produces a new state as output: $h_i = \lstm(h_{i-1}, \vy_i)$. Apart from the language model, an integral part of the caption generation process is selecting informative visual features from the videos to be fed to the language model (which is also the main contribution of this paper). A standard approach to this problem is to use~\emph{visual attention}. Mathematically, let $\ve\in\simplex{T}$ be a probability vector in the $T$-dimensional simplex; its $t$th dimension $\ve_t$ captures the probability that visual feature $\vx_t$ is useful for generating the $i$th word, typically given by:
\begin{equation}
\ve_t = \softmax\left(\att\left(h_{i-1}, \vx_t\right)\right),
\label{eq:2}
\end{equation}
where $\att$ is a suitable nonlinear attention function, usually chosen as 
\begin{equation}
\att(h_{i-1}, \vx_t)=\vw^{\mathrm T}\tanh\left(\mW_h h_{i-1} + \mW_x\vx_t + \bias\right).
\label{eq:attn}
\end{equation}
Here, $\bias$ is a learned bias, while $\mW_h$ and $\mW_x$ are learned matrices transforming the respective features into an attention space, in which they are linearly combined using the $\vw$ vector after passing through the nonlinear $\tanh$ function. The score $\ve$ is projected onto the simplex via the $\softmax$ operator in~\eqref{eq:2}, thereby generating a probability vector over the visual features. The visual features $\vx_t$ are linearly combined using weights $\ve_t$ to produce the attended visual feature.

\subsection{Spatio-Temporal (ST) Attention}
\label{sec:STspatial}
In this section, we present the Spatio-Temporal module of our attention framework. As may be noted, using multiple spatial (region-based) features (for every frame) introduces an additional degree of freedom in the visual domain (as against using only a single feature per frame), which needs to be attended to effectively. A straightforward way to extend the temporal attention in~\eqref{eq:2} to the spatio-temporal domain would be to ignore the spatial nature of these additional features and treat all $nT$ features as if they were the temporal features of the standard temporal attention model. However, given that each spatial feature could be noisy (i.e., containing features irrelevant or redundant to the end task), increasing the number of features to be attended may amplify the noise, thus diluting the attention paid to useful features. Further, there is temporal continuity in these features that should be incorporated in the method, for example to attend to spatially localized actions that span across frames. 

To circumvent such issues, we propose to compose the spatial and temporal attention one after the other. We explain the spatial aggregation in Section~\ref{sec:STspatialAttention}, then explain the subsequent temporal attention in Section~\ref{sec:STtemporalRanked}. Figure~\ref{fig:st_attn_model} illustrates our ST pipeline.

\subsubsection{ST Model: Spatial Attention}
\label{sec:STspatialAttention}
A direct way to implement spatial attention is to use~\eqref{eq:2} on each frame. That is, let $\ve_t^S$ denote the spatial attention for frame $t$:
\begin{equation}
    \ve^S_{tj} = \softmax\left(\att(h_{i-1}, \vx_{tj})\right), ~\text{where } \sum_{j=1}^n \ve^S_{tj}=1.  
\label{eq:st_attn}
\end{equation}
However, such a formulation makes no assumptions about the temporal relationships of the attended features from frame to frame. For example, when one needs to reason about the temporal evolution of video regions, say for generating the \emph{verb} part of a caption, a temporally-consistent spatial attention is preferred---we would like to attend to regions that contain the same entity over multiple frames. \emph{But how can we generate such consistent attention in a computationally inexpensive way?} We propose a simple way to achieve this by making some practical assumptions about the way the spatial regions are organized in the videos. Specifically, we assume these regions form a fixed non-overlapping grid (see the input CNN Features in Figure~\ref{fig:st_attn_model}), and each spatial feature summarizes the semantics in that grid location. Such an arrangement is a natural output of standard CNN pooling layers; e.g., the I3D model generates a $7\times7$ grid of spatio-temporal features. This grid is assumed to be consistent across all frames; as a result, when camera motion and scene changes are absent in the video, the features from the same grid cell across the frames are temporally consistent. However, when the camera moves or the scene changes, such an assumption no longer holds. 

We circumvent this problem by \emph{overestimating} the spatial attention region. Specifically, we propose a three-step process. First, we aggregate the spatial features at each grid cell across the temporal dimension, i.e., compute $\tilde{\vx}_j=\frac{1}{T}\sum_{t=1}^T \vx_{tj}$. Next, we use $\tilde{\vx}$ (which only contains $n$ features) in~\eqref{eq:st_attn} to compute spatial attention $\tilde{\ve}^S$. Finally, we replicate this attention to all frames: $\ve^S_t = \tilde{\ve}^S$ for all $t=1,2,\ldots, T$ (see Figure~\ref{fig:st_attn_model} middle block). 

Given that our proposed spatial attention is an approximate union of the attentions for individual frames, feature noise due to short scene changes or camera motion may be diluted when averaging the spatial features over all the frames. When training the framework end-to-end alongside the temporal ranked attention scheme (discussed in the next section), our overestimated attention will be guided to be correlated with regions in the video that exhibit dynamics, thereby pruning away non-action-related cues. Further, our heuristic also reduces the inference time linearly as the number of attentions to compute in this module is now independent of the number of frames in the sequence. Once the spatial attention $\ve_{tj}^S$ is computed, it is used to linearly average pool the spatial features for every frame (using~\eqref{eq:attn}), thus producing $T$ temporally-ordered features $\hat{\vx}_1, \hat{\vx}_2,\ldots, \hat{\vx}_T$ for the next module. 

\subsubsection{ST Model: Ranked Temporal Attention}
\label{sec:STtemporalRanked}
In this section, we detail our temporal pooling scheme, {\em ranked temporal attention} (also in Figure~\ref{fig:st_attn_model}). Using the spatially attended features $\hat{\vx}_1, \hat{\vx}_2,\ldots, \hat{\vx}_T$ produced by the spatial attention module described above, our goal is to capture the action dynamics in the input features. While there are several choices for modeling such dynamics popular in action recognition literature~\cite{carreira2017quo,yue2015beyond,feichtenhofer2017spatiotemporal}, we decided to use a model that is simple, effective, and lightweight. A standard approach is to use an LSTM for this task, but it is not guaranteed to capture the action dynamics unless it is trained with a suitable loss. 

To this end, we take inspiration from recent work on ranking-based dynamic feature pooling~\cite{bilen2016dynamic,fernando2015modeling,cherian2017generalized}. For temporally ordered inputs $\langle\hat{\vx}_1, \hat{\vx}_2,\ldots, \hat{\vx}_T\rangle$, these methods propose to compute a feature $\vw$ by solving the following rank-SVM formulation:
\begin{align}
    &\label{eq:rp1}\argmin_{\vw} \left[ \half\enorm{\vw}^2 + \lambda \sum_{t=1}^{T-1}   \softplus(\zeta_t)\right],\\
    &\text{where } \; \zeta_t =\langle\vw, \hat{\vx}_t\rangle + \beta - \langle\vw, \hat{\vx}_{t+1}\rangle,
    \label{eq:rankpool}
\end{align}
where $\lambda>0$ is a regularizer, and $\softplus(z)=\log(1+e^z)$ is a soft variant of the popular ReLU activation function. The rank pooling formulation seeks to find a direction \mbox{$\vw\in\reals{d}$} (same dimension as the input features) such that projecting the inputs to this direction will preserve their temporal order (with a margin of $\beta>0$), as enforced by the softplus function. Intuitively, the minimization encourages the projection of each frame's input feature, $\langle\vw, \hat{\vx}_{t+1}\rangle$, to be larger than the projection of the previous frame's input feature, $\langle\vw, \hat{\vx}_{t}\rangle$. Thus, the intuition is that this direction $\vw$, which lies within the input space, captures the temporal order (temporal dynamics), and can be used as an aggregated video feature for subsequent tasks. This has been found to be empirically useful in several recent works~\cite{bilen2016dynamic,cherian2017generalized}.

\begin{figure}[t]
    \centering
    \includegraphics[width=8cm,trim={1.5cm 6.3cm 7.5cm 3cm},clip]{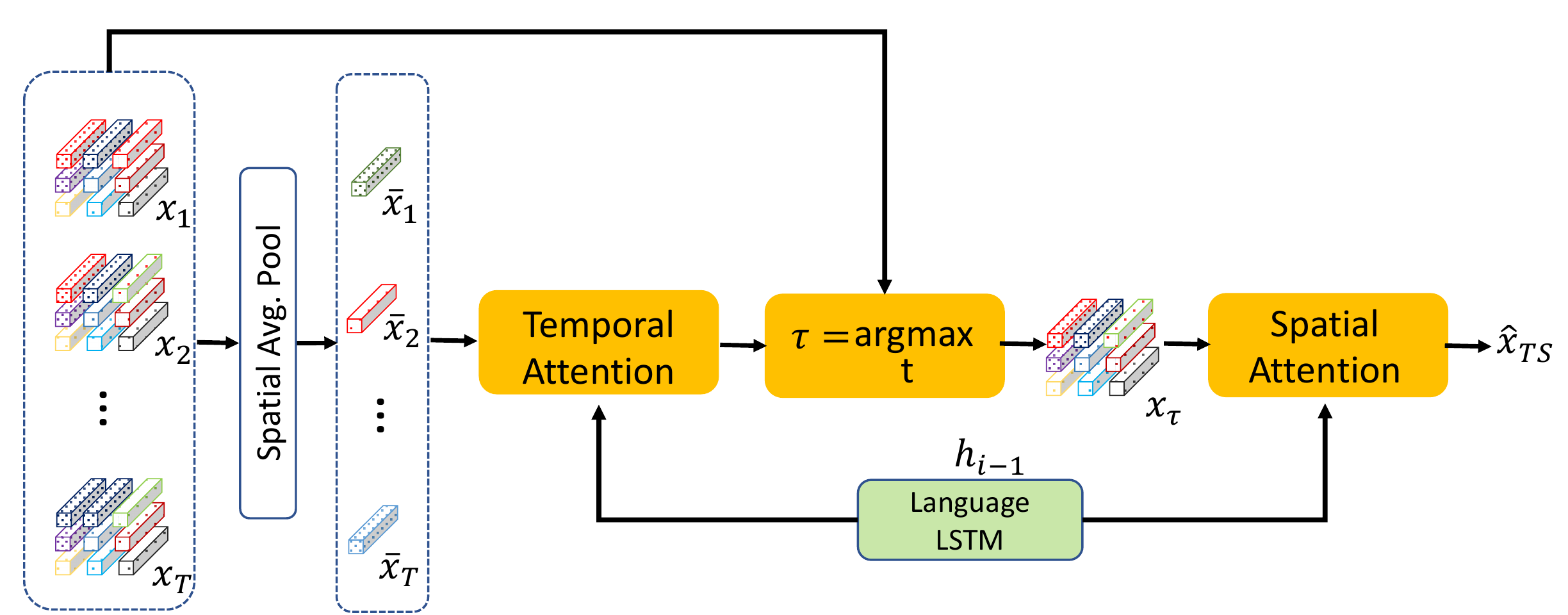}
    \caption{Our temporo-spatial (TS) attention module.}
    \label{fig:ts_attn_model}
\end{figure}
However, there is an important caveat for directly using rank pooling within a deep CNN framework: namely,~\eqref{eq:rp1} involves computing an $\argmin$ function, which is not differentiable. While there are workarounds for computing the derivative of this function~\cite{gould2016differentiating}, they lead to second-order gradients, which can be computationally expensive and may even be infeasible when the feature dimensionality is high. To circumvent this problem, we propose a simple scheme in this paper that we call \emph{ranked attention}. 

Our key idea is backed by the well known theoretical result that a recurrent neural network can approximate any algorithm (Turing machine)~\cite{siegelmann1992computational}. Motivated by this result, we propose to emulate the ranking SVM solution described above within an LSTM setting, such that it takes as input the sequence of features and produces a feature $\vw$ as output while also minimizng the softplus loss specified by~\eqref{eq:rp1}. Specifically, suppose the $\lstm$ is an abstract function~\cite{hochreiter1997long} parametrized by weights $\theta$. Then, using the above notation, we define our temporal pooling module (during training) as one that generates a representation $\hat{\vx}_{ST}$ by learning $\theta$ that optimizes the following loss:
\begin{align}
    &\label{eq:rlstm}\min_{\theta} \sum_{t} \softplus(\zeta_t),\\
 &\text{where }\quad \zeta_t = \langle \hat{\vx}_{ST}, \hat{\vx}_{t-1}\rangle + \beta - \langle\hat{\vx}_{ST}, \hat{\vx}_{t}\rangle,\\
 &\text{and }\quad \hat{\vx}_{ST} = \bigoplus_{t=1}^T \lstm_{\theta}(\hat{\vx}_t).
 \label{eq:rank_lstm}
\end{align}
Here, $\hat{x}_{ST}$ denotes the final output of the LSTM after it has seen all $T$ features. (The notation $\oplus$ denotes the sequential nature of inputting the features $\hat{x}_1, \ldots, \hat{x}_T$ to the LSTM, one frame at a time, while updating its internal state.) Intuitively, the formulation~\eqref{eq:rank_lstm} learns to produce a feature representation that preserves the temporal order of the input features; these features were output by our spatial attention model. Since the entire system is trained end-to-end, minimizing the softplus loss in turn trains the spatial attention to attend to temporally varying features, i.e., action dynamics. In~\eqref{eq:rank_lstm}, we avoid optimizing through $\argmin$ as in~\eqref{eq:rankpool}, instead optimizing the LSTM parameters $\theta$ alongside other STaTS parameters while respecting the order constraints.

\subsection{Temporo-Spatial (TS) Attention Model}
\label{sec:TS}
The ST attention model may help the system generate caption words for dynamic visual features (e.g., verbs), but attention to such temporal cues may not be necessary when generating words for the \emph{subject} or \emph{object} in a caption. For example, consider the sentence \emph{a boy is playing with a ball}. Here, the verb \emph{playing} may benefit from ST attention. However, using the ST attention framework for generating words such as \emph{boy} or \emph{ball} may be overkill and inefficient, so we need a more direct way to infer them.

To this end, we propose a separate attention-over-attention model, which we call {\em temporo-spatial attention.} In this model, we first use the standard temporal attention scheme described in~\eqref{eq:2}, then greedily select a single frame (or a few frames) to attend to (see Figure~\ref{fig:ts_attn_model}). Next, we apply spatial attention only to the features within these frames. Mathematically, suppose $\bar{\vx}_t$ represents a spatially agglomerated feature representation for frame $t$ (here $\bar{\vx}_t$ could be the average of all the spatial features for this frame, or a Max-Pooled vector). Our temporo-spatial (TS) attention is thus:
\begin{align}
\tau &= \arg\max_t \att(h_{i-1}, \bar{\vx}_t),\\
\ve_j^{TS} &= \att(h_{i-1}, \vx_{\tau j}), \quad \text{ where } \sum_j \ve_j^{TS} = 1.
\label{eq:ts_attn}
\end{align}
We define the {\em temporo-spatial attention} feature as: 
\begin{equation}
\hat{\vx}_{TS} = \sum_{j} \ve_j^{TS} \vx_{\tau j} \, .
\end{equation}
Note that while we write the frame selection via an $\arg\max$ function, we implement it via a $\softmax$ with a low temperature, as otherwise the model is non-differentiable.

\subsection{Spatio-Temporal and Temporo-Spatial Model}
\label{sec:STaTS}
For our full STaTS model, we combine the ST and TS models defined above via a further language attention-based weighting (see Figure~\ref{fig:overall_arch}). Let $\beta_1$ and $\beta_2$ be weight scalars: $\beta_1=w_{ST}\tanh(W_{ST}\hat{\vx}_{ST} + W_h h_{i-1})$ and $\beta_2=w_{TS}\tanh(W_{TS}\hat{\vx}_{TS} + W_h h_{i-1})$, where $W_{TS}, W_{ST}, w_{ST}, w_{TS}$ are learned parameters. Our STaTS model produces a combined feature representation:
\begin{equation}
\hat{\vx} = \tanh\left(\frac{\exp(\beta_1) \hat{\vx}_{ST} + \exp(\beta_2) \hat{\vx}_{TS}}{\exp(\beta_1) + \exp(\beta_2)}\right).
\label{eq:stats}
\end{equation}
This is another level of attention conditioned on the language state, which determines how much to attend to each attention branch (ST or TS) when generating the next caption word. 

\subsection{Model Training}
Our STaTS model is trained end-to-end using the ground truth video captions. A natural question in this regard is: what loss should we use? While softmax cross-entropy loss is the standard loss to consider, it is often argued that the cross-entropy may be only weakly correlated with the evaluation metrics we typically use on captions (such as METEOR or BLEU score). However, these metrics are non-differentiable and thus cannot be directly used. To this end, we follow~\cite{ranzato2015sequence,liu2017improved} to consider these metrics as reward functions in a reinforcement learning setup, and use policy gradients via the REINFORCE algorithm for optimizing against them. Specifically, following~\cite{ranzato2015sequence}, we first optimize our STaTS model to minimize the cross-entropy loss (for about 10 epochs), then subsequent iterations are optimized using a combination of cross-entropy loss and METEOR+BLEU rewards. We also use teacher forcing via scheduled sampling ~\cite{bengio2015scheduled} to reduce exposure bias when training the model.

\section{Experiments}
\label{sec:expts}
To validate the effectiveness of our STaTS architecture, we present experiments on the MSVD~\cite{chen2011collecting} and the MSR-VTT datasets~\cite{xu2016msr}, two popular benchmarks for video captioning. The MSVD dataset includes 1970 videos, split into 1200 videos for training, 100 for validation, and 670 for test, which is the recommended evaluation. Each video has about 40 ground truth (human-generated) captions, and 13,010 distinct words. MSR-VTT is has 10K training and 2990 test sequences and nearly 200,000 captions.

\subsection{Implementation and Evaluation}
As the primary contribution of this work is our spatio-temporal attention model, we mainly use two state-of-the-art CNN architectures for generating such features: (i) the Inflated 3D architecture (I3D) proposed in~\cite{carreira2017quo}, which has shown state-of-the-art performance on activty recognition benchmarks; and (ii) Faster R-CNN algorithm~\cite{ren2015faster} using a ResNet-101 architecture (FRCNN). The I3D features are generated for two modalities: (i) temporal chunks of 16 RGB frames at a temporal stride of 16, and (ii) temporal chunks of 16 optical flow frames at stride of 16. The I3D model implicitly uses the Inception-V3 architecture; we extract the spatial features from the ``Mixed\_5c'' layer of this network, which are $2\times 7\times 7\times 1024$ dimensional, which we reshape to $7\times 7\times 2048$, where the first two dimensions capture a $7 \times 7$ spatial grid. We use the same for the flow features. For the FRCNN features, we pass each frame (at a stride of 16) through a region-pooled ResNet-101 network~\cite{he2016deep}. We detect a fixed 10 bounding boxes per frame and extract features from the last fully-connected layer of the network, resulting in $10\times 2048$ spatial features. However, unlike the grid-structured I3D features, the RCNN features are region-pooled without any fixed grid. On the MSR-VTT dataset, we provide results using ResNet-152 features as well, to understand the differences in our performance due to feature type. Note that for either dataset, there is no standard feature type for comparing to prior methods; e.g., PickNet~\cite{chen2018less} uses ResNet-152, while DenseCap~\cite{shen2017weakly} uses C3D.

We use PyTorch software to implement our models. The CNN features are pre-computed and are embedded into 512-dimensions, while the words are embedded in 256 dimensions. We use single-head self-attention on the previously generated words (recall that the caption is generated sequentially, word by word) before combining them with the LSTM state for visual attention. We use an additive attention scheme with the query and key combined in an attention space of 128 dimensions~\cite{vaswani2017attention}. The models are trained using RMSprop algorithm with a learning rate of 0.0001. The training usually converges in about 20 epochs. We use a batch size of 32 for I3D or FRCNN features. The scheduled sampling uses a teacher forcing ratio of the form $\eta/\bigl(\eta+\exp(p/\eta)\bigr)$, where $\eta=24$ and $p$ is the epoch. To evaluate the performance of our models, we use BLEU4~\cite{papineni2002bleu}, METEOR~\cite{denkowski2014meteor}, ROUGE-L~\cite{lin2004rouge} and CIDEr~\cite{vedantam2015cider}. For fair comparisons with previous work, we compute scores using the code released on the Microsoft COCO evaluation server~\cite{chen2015microsoft}.

\begin{table}[]
    \centering
    \footnotesize\scalebox{0.85}{
    \begin{tabular}{l|c|c|c|c|c|c}
       Dataset & Scheme  & Feature &   CIDEr & BLEU4 & ROGUE & METEOR\\
    \hline
    \hline
     MSVD  &ST    & I3D &  0.742 & 0.502  & 0.68  & 0.325\\
        &TS  & I3D	 &  0.521 &	0.391 &	0.646 &	0.289\\
        &STaTS  & I3D &	\textbf{0.802} &	\textbf{0.526} &		\textbf{0.695} &	\textbf{0.335}\\
        \hline 
        &ST  &   FRCNN      &   0.686 &	0.477  &	0.69  &	0.33 \\
        &TS  & FRCNN        &	0.439 &	0.376 &		0.633 &	0.274\\
        &STaTS  & FRCNN     &   \textbf{0.709} &	\textbf{0.492} &	\textbf{0.68}  &	\textbf{0.319}\\
     \hline
     \hline
    MSR-VTT&ST  &   I3D       &   0.429 &	0.397  &	0.600  &	0.271 \\
        &TS  & I3D         &	0.427 &	0.380 &		0.595 &	0.273\\
        &STaTS  & I3D      &   \textbf{0.434} &	\textbf{0.401} &	\textbf{0.604}  &	\textbf{0.275}\\
    \end{tabular}}
    \caption{Combinations our method on the MSVD and MSR-VTT datasets using the I3D (RGB) and Faster R-CNN features.}
    \label{tab:msvd-i3d-spatial}
\end{table}

\begin{figure*}[ht]
    \centering
    \includegraphics[width=0.9\linewidth]{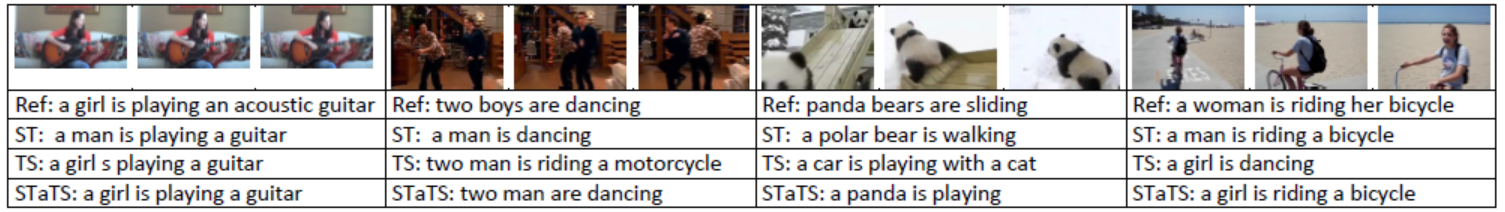}
  \caption{Qualitative results using our attention model.}
    \label{fig:my_label}
      \vspace*{-0.3cm}
\end{figure*}

\begin{figure}[ht]
    \centering
    \includegraphics[width=0.8\linewidth]{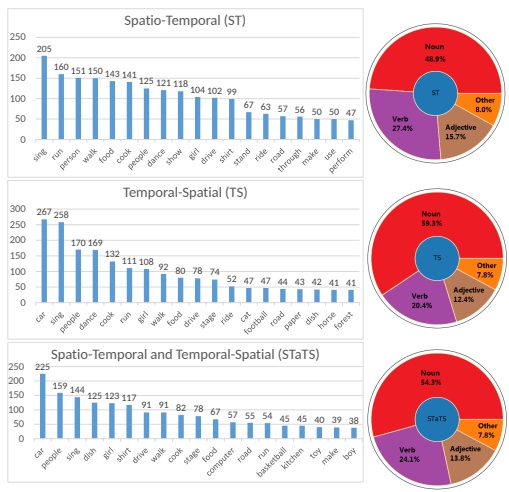}
  \caption{Words distribution analysis for generated captions in MSR-VTT testing set.}
    \label{fig:bar_and_pie}
\end{figure}

\begin{figure}[ht]
    \centering
    \includegraphics[width=8cm]{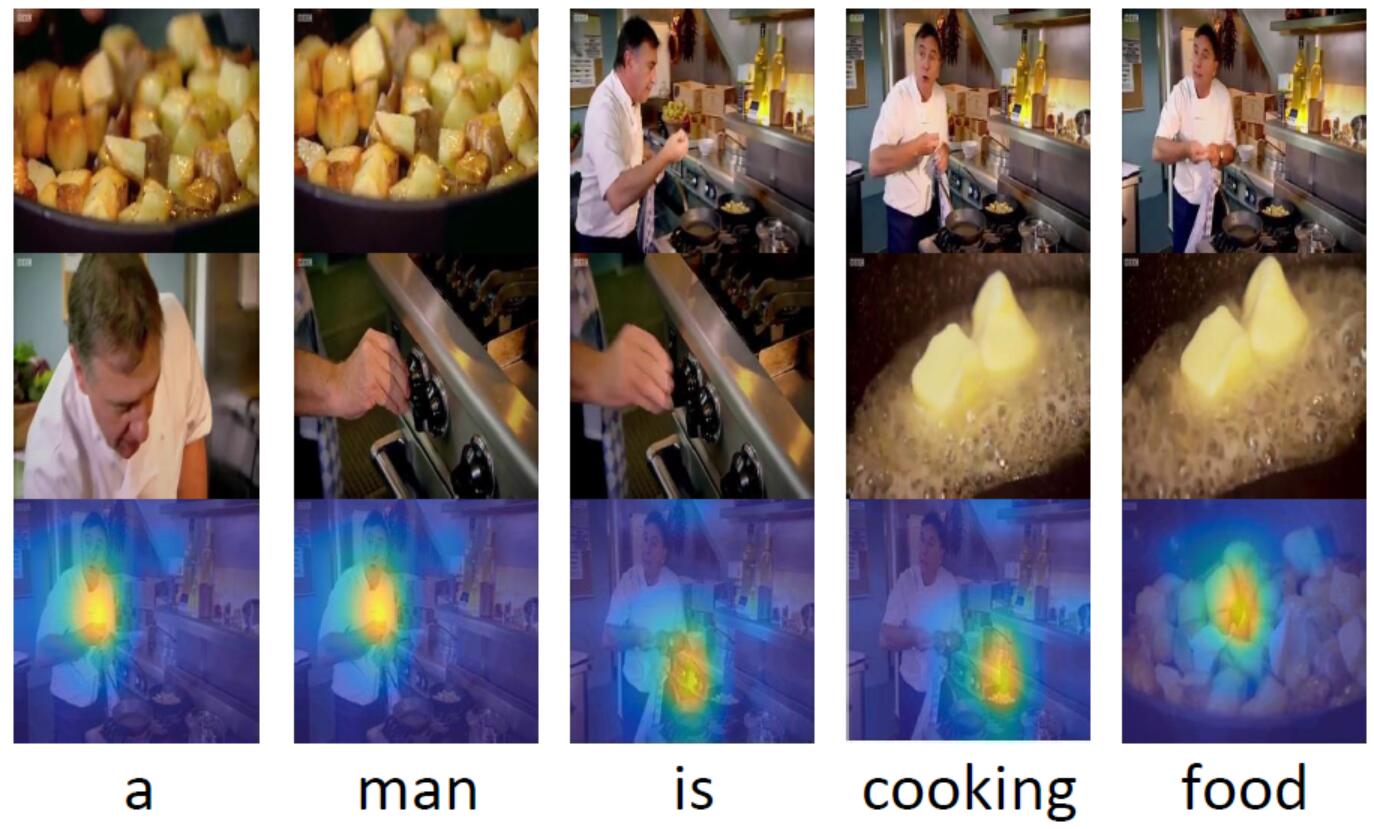}
  \caption{Attention Visualization. The 10 frames in the first two rows show the temporal sequence of the video. The 3rd row shows the frames selected by our TS model for each word in the generated caption, overlaid with its corresponding spatial attention map.} 
  \vspace*{-0.5cm}
    \label{fig:good1}
\end{figure}

\begin{table}[]
    \centering
    \footnotesize\scalebox{0.9}{
    \begin{tabular}{l|c|c|c|c}
        Scheme   &   CIDEr & BLEU4  & ROGUE & METEOR\\
    \hline
        Mean Pool& 0.389 & 0.362 & 0.580 & 0.263\\
        LSTM     &0.385 & 0.347 & 0.578 & 0.261\\
        Mean + LSTM &	0.388 &	0.364 &	0.575	& 0.259\\
        Temp Att & 0.382 & 0.368 & 0.580 & 0.258\\
        Mean + Temp Att &	0.385 &	0.368	& 0.58	& 0.26\\
        \hline
        Ranked Att (ours) & 0.387 & 0.376 & 0.589 & 0.264\\
        Mean + Ranked Att (ours) & \textbf{0.404} & \textbf{0.376} & \textbf{0.592} & \textbf{0.268}\\
    \end{tabular}}
    \caption{Study on the benefits in using Ranked Attention. The results are on the MSR-VTT dataset using the I3D (RGB) features.}
    \label{tab:msrvtt-scheme}
\end{table}

\begin{table}[]
    \centering
    \footnotesize\scalebox{0.9}{
    \begin{tabular}{l|c|c|c|c}
        Scheme &     CIDEr & BLEU4 & ROGUE & METEOR\\
    \hline
        PickNet~\cite{chen2018less}& 0.765 & 0.523 & 0.696 & 0.333\\
        $M^3$~\cite{mmm}  &  N/A & 0.520 &  N/A & 0.321\\
        LSTM-LS~\cite{liu2017video} & N/A & 0.511 & N/A & 0.326\\
        MA-LSTM~\cite{xu2017learning} & 0.704 & 0.523  & N/A & 0.336\\
        MAM-RNN~\cite{li2017mam} & 0.539 & 0.413 & 0.688 & 0.322\\
        RecNet~\cite{RNV} & 0.803 & 0.523 &  0.698 & 0.341 \\
        GRU-EVE~\cite{Aafaq_2019_CVPR} &   0.781 & 0.479  & \textbf{0.715} & \textbf{0.350}\\
        \hline
        STaTS(FR+FL) &   0.747 &    0.495 &	0.694 &	0.334\\
        STaTS (I3D+FL) & \textbf{0.835} &	\textbf{0.548} &		0.711 &	\textbf{0.350} \\

    \end{tabular}}
    \caption{Comparisons to the state of the art on MSVD dataset. FR standds for FRCNN models, I3D and FL stands for the I3D RGB and optical flow models respectively.}
    \label{tab:msvd-combination}
\end{table}

\begin{table}[]
    \centering
    \footnotesize\scalebox{0.9}{
    \begin{tabular}{l|c|c|c|c}
        Scheme &     CIDEr & BLEU4 & ROGUE & METEOR\\
        \hline
        Dense-Cap~\cite{shen2017weakly}  & \textbf{0.489} & 0.414 & 0.611 & 0.283\\
        PickNet~\cite{chen2018less} & 0.441 & 0.413 & 0.598 & 0.277\\
        OA-BTG (R200)~\cite{Zhang_2019_CVPR} & 0.469 & 0.414 & -- & 0.282 \\
        M$^3$-VC~\cite{mmm} & -- & 0.381 & -- & 0.266\\
        GRU-EVE (C3D+IVR2)~\cite{Aafaq_2019_CVPR} &  0.481  & 0.383 & 0.607 & \textbf{0.284}\\
        RecNet~\cite{RNV} & 0.427 & 0.391 & 0.593 & 0.266 \\
        \hline
        STaTS (R152) & {0.445} &	0.392 &	0.597 &	0.279\\
        STaTS (R152+C3D) & {0.465} &	0.416 &		\textbf{0.615} &	\textbf{0.284}\\
        \hline
        STaTS (I3D) &   0.434 &   0.401 &	0.604 &	0.275\\
        STaTS (I3D+FL) &   0.438 &   0.410 &	0.611 &	0.276\\	
        STaTS (I3D+FL+C) & {0.451} &	\textbf{0.417} &		{0.612} &	{0.280}
    \end{tabular}}
    \caption{Comparisons to the state of the art on MSR-VTT dataset. I3D and FL stand for the I3D RGB and optical flow models, respectively, while C stands for using the class annotations supplied with the dataset during training (as is also used by other methods).}
    \label{tab:msrvtt-soa}
\end{table}
\subsection{Results}
In the following, we first conduct an ablation study of the various components in our framework.
\noindent\textbf{ST Spatial Attention:} Table~\ref{tab:msvd-i3d-spatial} shows the performance on MSVD and MSR-VTT datasets using I3D and FRCNN features with various attention schemes. We show the performance when using only our spatio-temporal (ST) model, only temporo-spatial (TS), and our combined STaTS model. TS is usually the weakest model, likely due to its greedy attention scheme. The table shows that there is significant synergy between the ST and TS models as substantiated on both the datasets. The table also compares our approximate ST attention and grid-based feature organization (using I3D features) against the alternative of attending to different image regions per frame (using FRCNN features). This comparison (ST and TS in the first two meta-rows of Table~\ref{tab:msvd-i3d-spatial}) shows that our heuristic performs significantly better than FRCNN on CIDER and BLEU4, which are measures capturing the exact match of parts of the generated caption with the ground truth. Also, comparing the full STaTS model using I3D and FRCNN shows that our model is substantially better (0.802 vs. 0.709 on CIDER).

\noindent\textbf{Ranked Attention:} In Table~\ref{tab:msrvtt-scheme}, we demonstrate the benefits of our ranked temporal attention scheme versus several other plausible choices on the MSR-VTT dataset using the I3D RGB features. We compare to: (i) using mean pooling of the spatially-pooled temporal features, (ii) using an LSTM, (iii) combining LSTM with average pooling, (iv) temporally attending over all spatio-temporal features (no ST-attention nor ranked attention), and (v) using average pooling of spatial features and then temporal pooling of them. We see that while the ranked attention by itself is not significantly better than other choices, combining ranking with average pooling demonstrates the best performance. This is not surprising, given that the ranked attention considers only the ordering of the features, but discards features that are invariant to temporal permutation (features which are captured by mean pooling). We use the combination of mean-pooling + ranked attention in our subsequent model.

\noindent\textbf{Qualitative Comparisons:} Figure~\ref{fig:my_label} shows improvements provided by each module. We find that the ST model captures more of action-related cues and provide caption verbs, while the TS model better captures the appearances predicting the right nouns. The STaTS module absorbs the benefits from both ST and TS, yielding the best video captioning. To back up these qualitative observations, Figure~\ref{fig:bar_and_pie} provides more insight into how different attention modules affect the resulting caption. In the bar chart, we sort all the words from the generated captions for the testing set of MSR-VTT according to their frequency. First, we remove the top 5 most frequent words from each chart (such as ``man'' and ``woman''). Each bar chart shows the top 20 verbs and nouns, from which it can be seen that the ST module generates more verbs (13 verbs out of 20) while the TS module generates more nouns (12 nouns out of 20). A similar phenomenon is shown in the adjacent pie charts, which indicate the total percentage of verbs, nouns, and adjectives in the generated captions. Notably, the ST model generates nearly 27\% verbs (8\% higher than the TS model), while the TS model generates 59\% nouns (10\% higher than the ST model), demonstrating their complementary nature. The combination of ST and TS, the STaTS module, provides a balance between the two. In Figure~\ref{fig:good1}, we visualize an example of how STaTS attention is localized spatially and temporally in the sequence (more examples in the supplementary materials). The first two rows illustrate the sequence of events in the video. The third row visualizes the attention. For each word in the generated caption (fourth row), we chose the frame with highest temporal attention and overlaid the respective spatial attention.

\textbf{Comparisons to the State of the Art:} In Table~\ref{tab:msvd-combination}, we show the results of our STaTS method with various feature combinations and compare it against state-of-the-art methods on the MSVD dataset. Our model fares better by more than 3.5\% on the CIDEr and by 2\% on BLEU4 than the next best method (RecNet~\cite{RNV}). In Table~\ref{tab:msrvtt-soa}, we provide comparisons on the MSR-VTT dataset. We outperform several recently proposed methods. Specifically, we outperform RecNet on all four metrics, while  outperforming more recent GRU-EVE~\cite{Aafaq_2019_CVPR} and OA-BTG~\cite{Zhang_2019_CVPR} on most metrics. Note that these are powerful deep models that combine visual saliency with dynamics learning, and our results clearly demonstrate the superiority of our approach. 
\vspace*{-0.2cm}
\section{Conclusions}
\label{sec:conclude}
We proposed novel attention models for video caption generation combining spatio-temporal and temporo-spatial (STaTS) attention. We also presented ranked temporal pooling using an LSTM that emulates a rank-SVM. Our method can be seen as  stage-wise attention, in which spatial and temporal cues are explored hierarchically. Our scheme yields state-of-the-art results on two benchmark datasets.

{\small
\bibliographystyle{ieee}
\bibliography{stats_bib}
}
\section*{Qualitative results}
\begin{figure*}[ht]
    \centering
    \includegraphics[width=0.70\linewidth]{figs/good5.jpg}
    \includegraphics[width=0.80\linewidth]{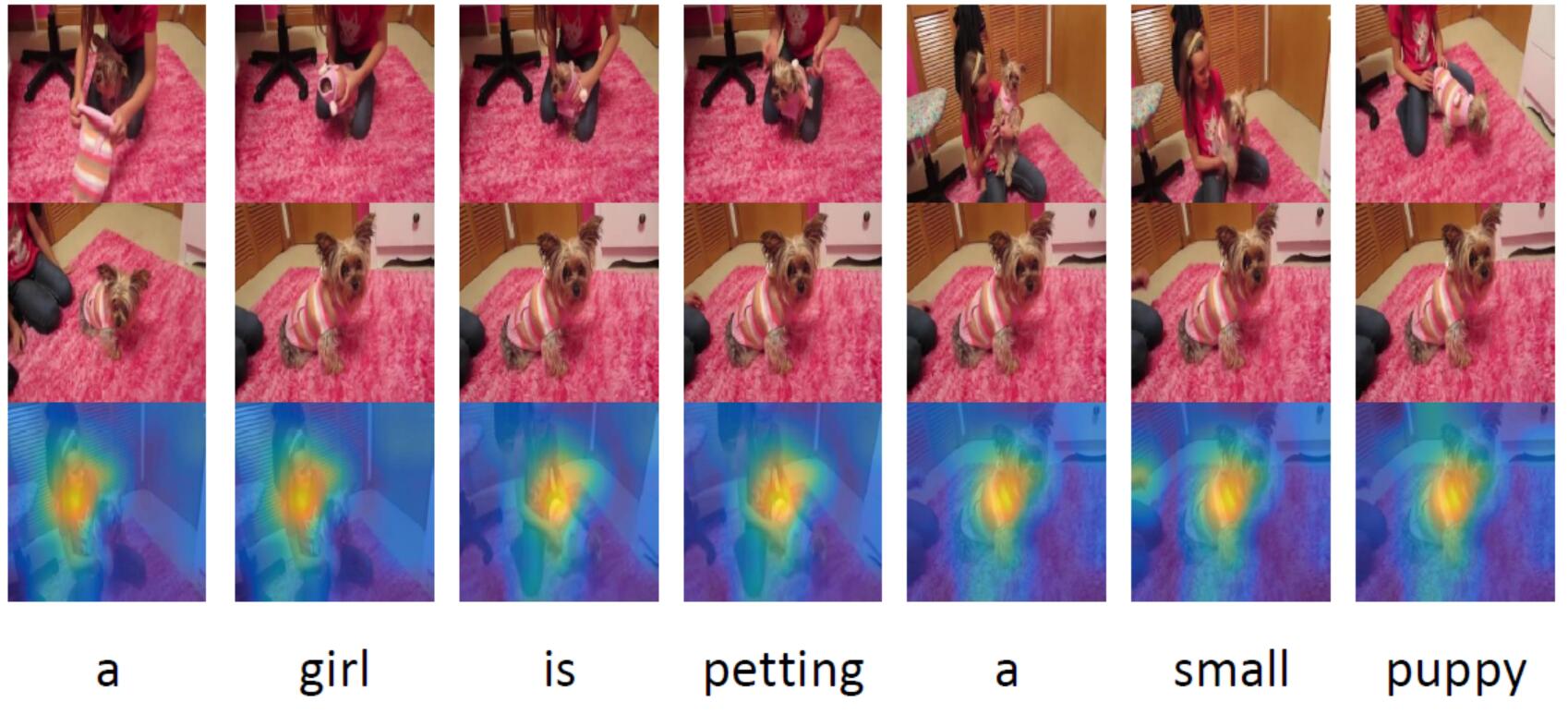}
    \includegraphics[width=0.8\linewidth]{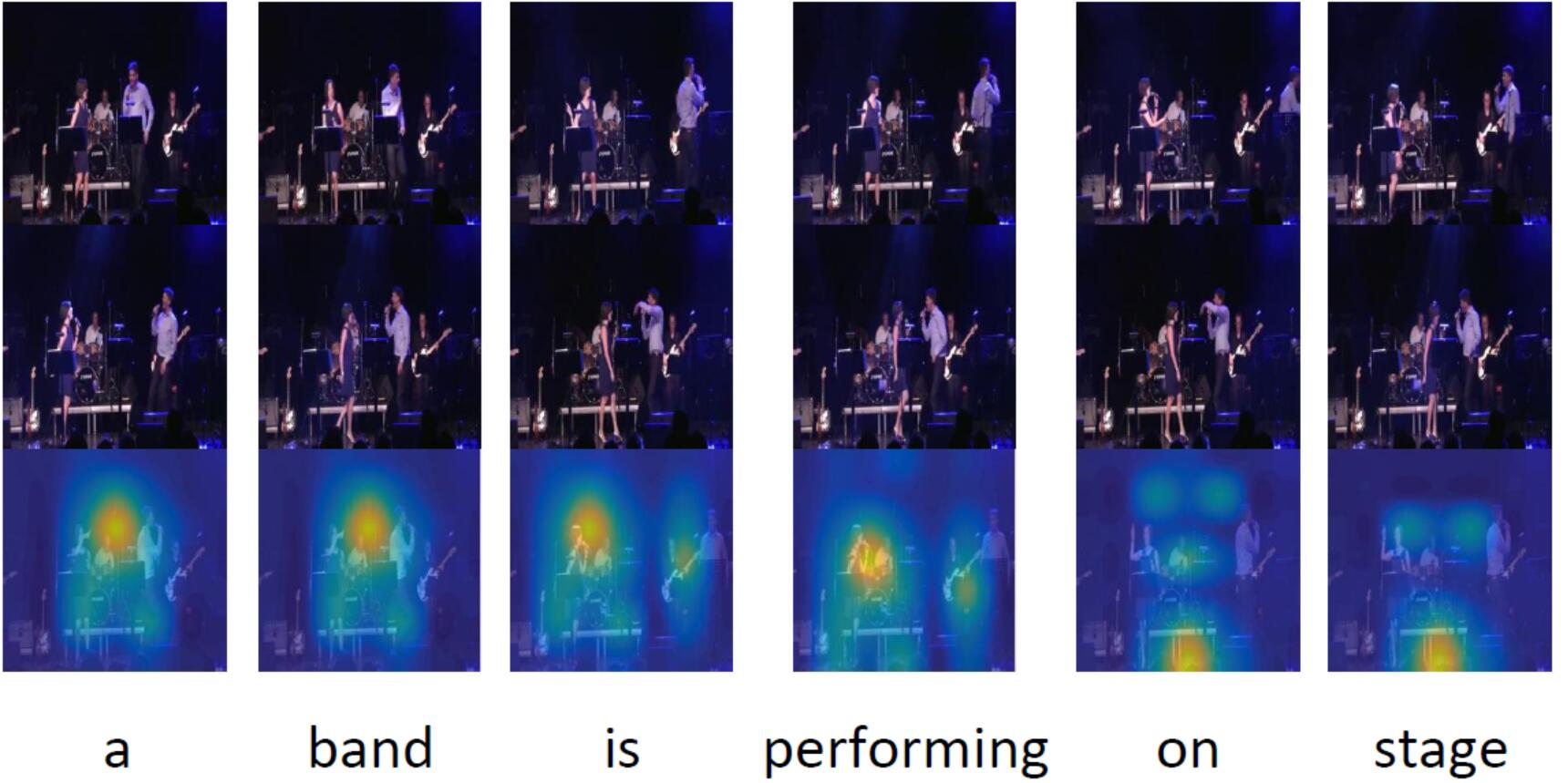}
  \caption{Attention Visualization. The frames in the first two rows show the temporal sequence of the video. The 3rd row shows the frames selected by our TS model for each word in the generated caption, overlaid with its corresponding spatial attention map.}
    \label{fig:good}
\end{figure*}

\begin{figure*}[]
    \centering
    \includegraphics[width=0.8\linewidth]{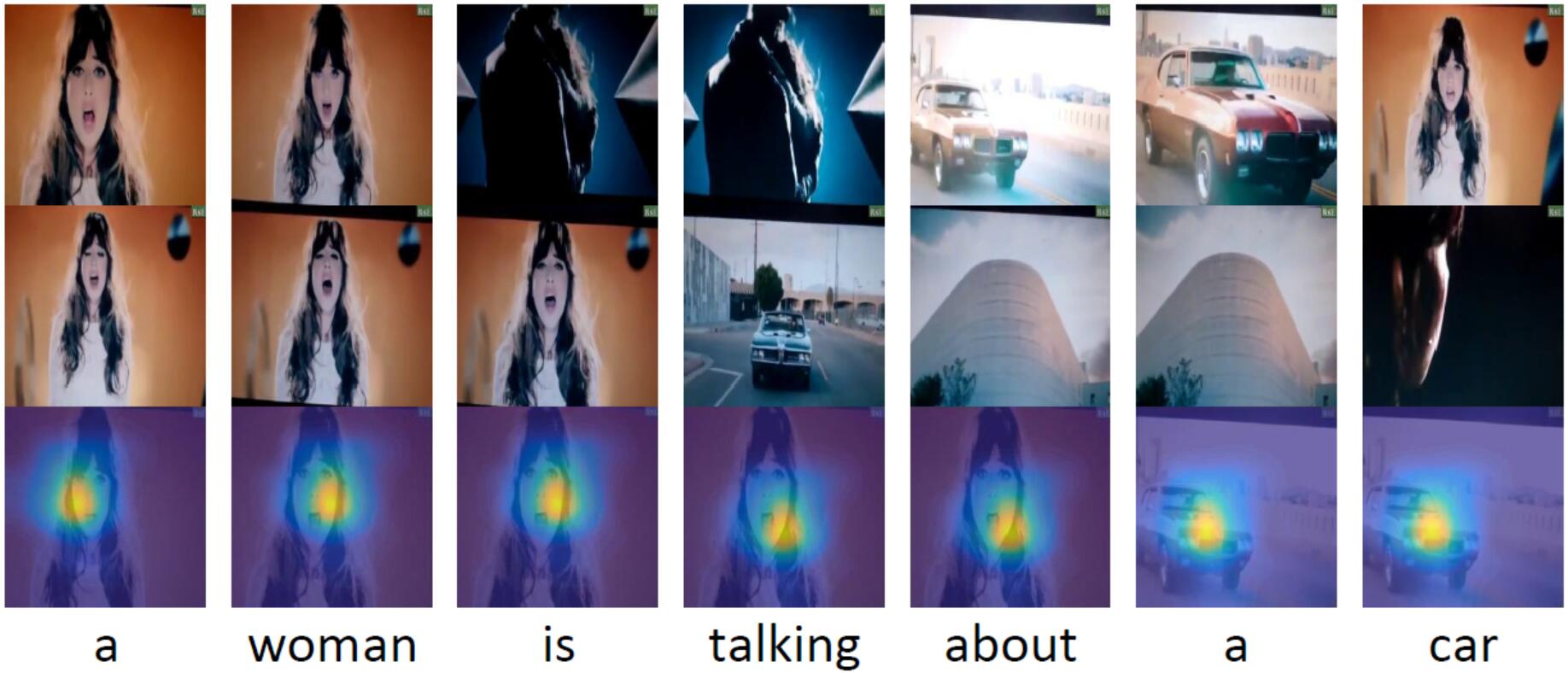}
  \caption{Failure case: the system fails to recognize the sound.  The 14 frames in the first two rows show the temporal sequence of the video. The 3rd row shows the frames selected by our TS model for each word in the generated caption, overlaid with its corresponding spatial attention map.}
    \label{fig:bad}
\end{figure*}

In Table~\ref{tab:1}, we provide examples of video captions generated by our scheme and the human generated captions; for the latter, we randomly selected one caption (out of 20) to show for the respective video. Our provided results are using the STaTS model with I3D features on the MSR-VTT dataset.
In Figures~\ref{fig:good} and \ref{fig:bad}, we show qualitative attentions on the respective video frames, the former showing examples when our captions are very similar to human captions, and the latter showing some failure cases. In Figure~\ref{fig:my_label}, we show additional results of our ST, TS, and STaTS attention.

\begin{table*}[ht]
    \centering
    \begin{tabular}{l|p{7cm}|p{7cm}}
    Test id\# & Reference caption & Generated caption\\
    \hline
         1 (7517) & a woman is demonstrating various features of a car  &  a car is being shown \\
         \hline
         2 (9987)  & a finger goes around the corners of a piece of paper & a person is folding a piece of paper\\
         \hline
         3 (7030) & a ballroom dance class & a group of people are dancing \\
         \hline
         4 (7519) & optimus prime voice is used briefly during video game play & a man is playing a video game\\
         \hline
         5 (7518) & a game character is floating in space & a minecraft character is talking\\
         \hline
         6 (8697) & a boy is sitting on a chair outside he is being recorded while he sings and plays the guitar & a man is singing a song\\
         \hline
         7 (8696) & a guy swims in blue goggles & a woman is swimming in the water\\
         \hline
         8 (7886) & a man is demonstrating how to slice a potato thinly using a knife and a cutting board & a man is cutting potatoes\\
         \hline
         9 (9525) & a chef slices up a fish & a woman is showing how to make a dish\\
         \hline
         10 (8168) & a guy is playing golf & a man is talking about a dog\\
         \hline
         11 (8765) & a guy opens a box for a toy car & a man opens a box\\
         \hline
         12 (9405) & red balloons containing small gifts dropping to the people of the city & a group of people are playing a rocket\\
    \end{tabular}
    \caption{Captions generated our STaTS model and the corresponding human generated caption for the video. The video id from the MSR-VTT dataset is also shown. }
    \label{tab:1}
\end{table*}
\begin{figure*}[ht]
    \centering
    \includegraphics[width=0.9\linewidth]{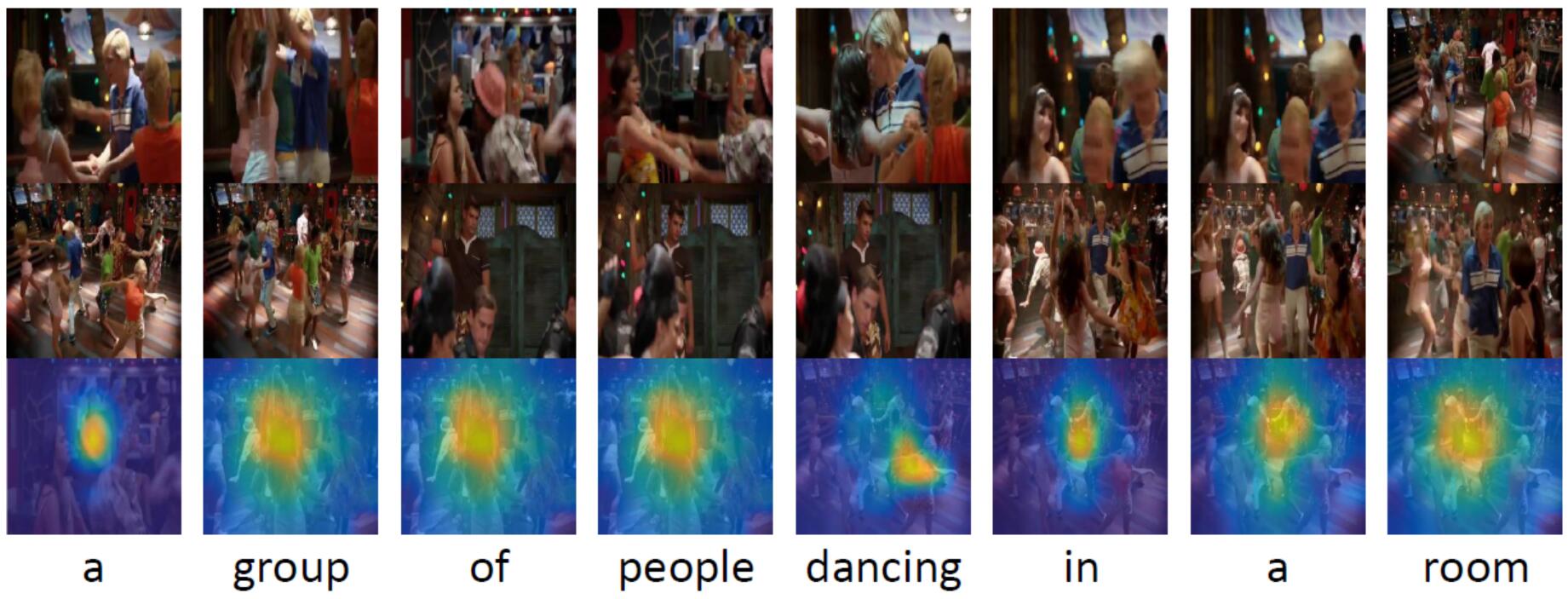}
  \caption{Attention Visualization. The frames in the first two rows (of each sub figure) show the temporal sequence of the video. The 3rd row shows the frames selected by our TS model for each word in the generated caption, overlaid with its corresponding spatial attention map.}
    \label{fig:good1}
\end{figure*}
\begin{figure*}[]
    \centering
    \includegraphics[width=0.8\linewidth,trim={22.89cm 0cm 0cm 0cm},clip]{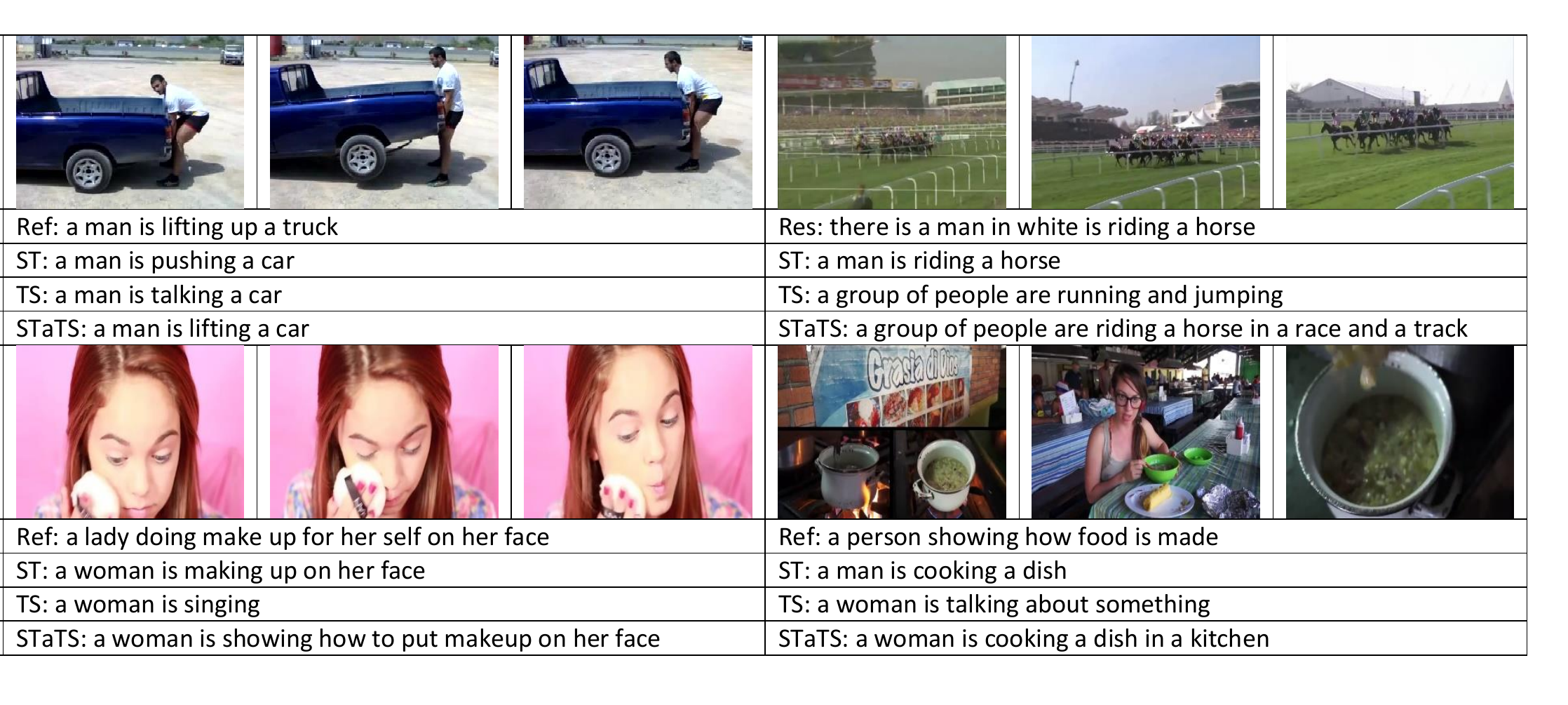}
    \hspace*{-0.5cm}\includegraphics[width=0.8\linewidth,trim={0cm 0cm 22.69cm 0cm},clip]{figs/example_cap.pdf}
  \caption{Qualitative results using our attention model.}
    \label{fig:my_label}
\end{figure*}

\end{document}